%% file: paper.tex
\documentclass[letterpaper, 10 pt, conference]{ieeeconf}  % Comment this line out if you need a4paper

\IEEEoverridecommandlockouts                              % This command is only needed if 
\title{\LARGE \bf
CA-W3D: Leveraging Context-Aware Knowledge for Weakly Supervised Monocular 3D Detection
}

\author{Chupeng Liu, Runkai Zhao$^{\dagger }$, Weidong Cai% <-this % stops a space
\thanks{$\dagger$ Corresponding author}
\thanks{Chupeng Liu, Runkai Zhao, Weidong Cai are with School of Computer Science, Faculty of Engineering, The University of Sydney, NSW, Australia.
Email: {\tt\small cliu6713@uni.sydney.edu.au; rzha9419@uni.sydney.edu.au; tom.cai@sydney.edu.au}}%
}

\usepackage{multirow}    % 支持 multirow 合并单元格
\usepackage{graphicx}    % 插入图表
\usepackage{amsmath}     % 数学公式支持
\usepackage{subcaption}  % 支持子表格功能
\usepackage{xcolor}
\usepackage[
    colorlinks=true,      % 启用颜色链接
    citecolor=blue,        % 设置引用颜色
    urlcolor=cyan         % 设置 URL 颜色
]{hyperref}
\usepackage{cite}      % 自动格式化引用
\usepackage{algorithm}
\usepackage{algorithmic}
\usepackage{amssymb}
\usepackage{booktabs}
\usepackage{pdfpages}
\usepackage{graphicx}  % 用于插入图像
\usepackage{lipsum}    % 用于生成示例文本
\usepackage{float}     % 强制图片位置
\usepackage{wrapfig}
\usepackage{listings}
\usepackage{xcolor}
\usepackage{svg}
\usepackage[utf8]{inputenc}
\usepackage{makecell}     % 支持 makecell

\definecolor{commentblue}{rgb}{0,0,1} % 设置蓝色
\begin{document}

\maketitle
\thispagestyle{empty}
\pagestyle{empty}

%%%%%%%%%%%%%%%%%%%%%%%%%%%%%%%%%%%%%%%%%%%%%%%%%%%%%%%%%%%%%%%%%%%%%%%%%%%%%%%%
\begin{abstract}
\input{abstract}
\end{abstract}

%%%%%%%%%%%%%%%%%%%%%%%%%%%%%%%%%%%%%%%%%%%%%%%%%%%%%%%%%%%%%%%%%%%%%%%%%%%%%%%%
\input{Introduction.tex}

%%%%%%%%%%%%%%%%%%%%%%%%%%%%%%%%%%%%%%%%%%%%%%%%%%%%%%%%%%%%%%%%%%%%%%%%%%%%%%%%
\input{Related_Works.tex}

%%%%%%%%%%%%%%%%%%%%%%%%%%%%%%%%%%%%%%%%%%%%%%%%%%%%%%%%%%%%%%%%%%%%%%%%%%%%%%%%
\input{Methodology_2.tex}

%%%%%%%%%%%%%%%%%%%%%%%%%%%%%%%%%%%%%%%%%%%%%%%%%%%%%%%%%%%%%%%%%%%%%%%%%%%%%%%%
\input{Experiments_and_Results.tex}

%%%%%%%%%%%%%%%%%%%%%%%%%%%%%%%%%%%%%%%%%%%%%%%%%%%%%%%%%%%%%%%%%%%%%%%%%%%%%%%%
\input{Conclusion_and_Discussion.tex}

%%%%%%%%%%%%%%%%%%%%%%%%%%%%%%%%%%%%%%%%%%%%%%%%%%%%%%%%%%%%%%%%%%%%%%%%%%%%%%%%

\bibliographystyle{IEEEtran} % 使用 IEEEtran.bst
\bibliography{main}

\end{document}

%% file: abstract.tex
Weakly supervised monocular 3D detection, while less annotation-intensive, often struggles to capture the global context required for reliable 3D reasoning. Conventional label-efficient methods focus on object-centric features, neglecting contextual semantic relationships that are critical in complex scenes. In this work, we propose a \underline{\textit{C}}ontext-\underline{\textit{A}}ware \underline{\textit{W}}eak Supervision for Monocular \underline{\textit{3D}} object detection, namely \underline{\textit{CA-W3D}}, to address this limitation in a two-stage training paradigm. Specifically, we first introduce a pre-training stage employing \textit{Region-wise Object Contrastive Matching} (\textit{ROCM}), which aligns regional object embeddings derived from a trainable monocular 3D encoder and a frozen open-vocabulary 2D visual grounding model. This alignment encourages the monocular encoder to discriminate scene-specific attributes and acquire richer contextual knowledge. In the second stage, we incorporate a pseudo-label training process with a \textit{Dual-to-One Distillation} (\textit{D2OD}) mechanism, which effectively transfers contextual priors into the monocular encoder while preserving spatial fidelity and maintaining computational efficiency during inference. Extensive experiments conducted on the public KITTI benchmark demonstrate the effectiveness of our approach, surpassing the SoTA method over all metrics, highlighting the importance of contextual-aware knowledge in weakly-supervised monocular 3D detection. For implementation details: \href{https://github.com/AustinLCP/CAW3D.git}{CAW3D}
\vspace{5pt}

%% file: Introduction.tex
\section{INTRODUCTION}

% weakly supervised monouclar 3D detection
Perceiving 3D objects from a single-view image has garnered increasing research attention as a cost-effective alternative to relying on expensive depth-sensing sensors such as RGB-D cameras \cite{xia2024rgbd, pesavento2024anim, huang2024matchu} and LiDAR \cite{lang2019pointpillars, zhou2017voxelnetendtoendlearningpoint, zhao2024advancements, yu2024unleashing, yu2024future}. Monocular 3D object detection lacks explicit depth information, making it heavily reliant on high-cost 3D annotations during training to encode spatial cues. To solve this challenge, label-efficient training strategies are recently developed, leveraging pseudo-3D label generation \cite{zhang2023odm3d, weng2019monocular, wang_plumenet_2021, you_pseudo-lidar_2020, wang_pseudo-lidar_2019, qian_end--end_2020} or geometry constraint-based supervision \cite{leonardis_weakly_2025, jiang_weakly_2024, zhang_decoupled_nodate, peng_weakm3d_2022} to guide 3D object feature extraction. Despite these advancements, such hand-crafted supervisions predominantly focus on the local regions around objects, neglecting higher-level semantic cues. This narrow emphasis limits the ability of monocular encoder to infer contextual relationships within complex scenes. As a result, weakly supervised monocular models struggle to capture the broader scene context necessary for global 3D reasoning. This raises an intriguing research question: \textit{how can a monocular encoder preserve essential spatial awareness while simultaneously acquiring the context-aware understanding under weak supervision?}
\begin{figure}[t]
  \centering
    \includegraphics[width=\columnwidth]{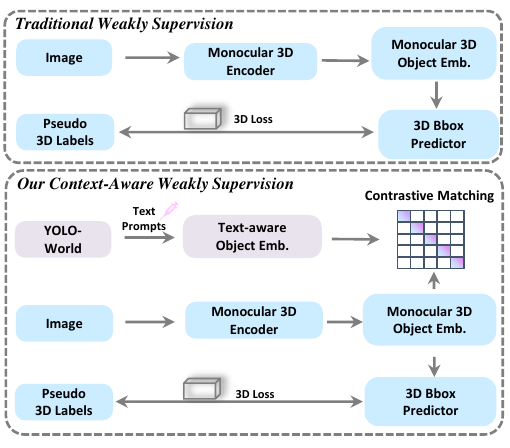} % 使图像与单栏对齐
  \caption{Weakly supervised monocular 3D object (Mono3D-Obj) detection relies solely on pseudo 3D labels to supervise the local object-specific features. Our proposed \textit{Context-Aware Weakly Supervision Mono3D-Obj (CA-W3D)} further guides monocular feature learning with context-aware global knowledge derived from an pre-trained open-vocabulary grounding model.}
  \label{fig:dessert_fig}
  \vspace{-15pt}
\end{figure}

\begin{figure}[t]
  \centering
    \includegraphics[width=0.9\columnwidth]{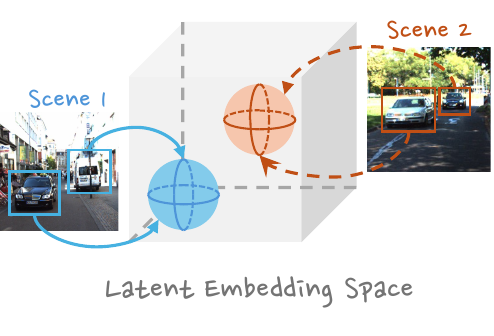} % 使图像与单栏对齐
  \caption{\textbf{Objects are embedded distantly in the latent space across scenes due to contextual variations.} Scenes 1 and 2 contain objects from the same class, such as cars, but appear in different contextual scenarios, leading to distinct embeddings in the latent space. The alignment process groups co-occurring objects more closely within the embedding space, enabling the monocular 3D encoder to capture higher-level scene semantics and global reasoning.}
  \label{fig:latent_embed_space}
  \vspace{-15pt}
\end{figure}

% Open 
A straightforward method to capture global contextual cues is merely scaling up by expanding model size or incorporating additional scene-level module \cite{oquab2024dinov2learningrobustvisual, singh2024smartmask}. These lead to increasing computational expense and are infeasible for real-world monocular 3D deployments. In light of these constraints, we explore multi-modal image-text alignment, a promising avenue in zero-shot detection and open-vocabulary recognition, as a more efficient means of infusing broader context into monocular image learning. Notably, CLIP \cite{radford2021learningtransferablevisualmodels} demonstrates strong capabilities in bridging visual and textual embeddings, mapping semantically related categories, such as “cat” and “elephant” under the “animals” concept, into a shared latent space. This alignment allows the model to generalize to unseen objects like “rabbit” by leveraging high-level category associations. Inspired by this principle, we hypothesize that objects within the same scene exhibit stronger semantic similarity than those across different scenes. As shown in Fig. \ref{fig:latent_embed_space}, by associating these co-occurrence objects more closely in the latent space, the monocular image encoder gains a high-level interpretation of the overall scene composition and semantics.

Building on this insight, we intend to introduce this contrastive learning pattern into a weakly supervised monocular training. As shown in Fig. \ref{fig:dessert_fig}, compared to traditional weakly supervision, our monocular 3D encoder is encouraged to recognize scene-specific object groupings while differentiating inter-scene variability without additional computational overhead. Specifically, we propose a novel \underline{\textbf{C}}ontext-\underline{\textbf{A}}ware \underline{\textbf{W}}eak Supervision for Mono\underline{\textbf{3D}}-Obj, namely \underline{\textbf{CA-W3D}}, which first introduces a pre-training stage that precedes pseudo-3D label training, enabling the monocular encoder to internalize scene-specific characteristics. Specifically, we propose \textbf{R}egion-wise \textbf{O}bject \textbf{C}ontrastive \textbf{M}atching (\textbf{ROCM}), aligning monocular object embedding with multi-modal object embeddings drawn from a frozen open-vocabulary 2D visual grounding model. During this warm-up stage, the encoder is encouraged to discriminate scene-dependent attributes, thereby acquiring richer contextual knowledge. To avoid latent feature conflict, we further decouple this learned context from spatial-awareness features, as the latter heavily depends on low-level texture cues. Additionally, we introduce a \textbf{D}ual-\textbf{to}-\textbf{O}ne \textbf{D}istillation (\textbf{D2OD}) mechanism in pseudo-3D label training, allowing to effectively transfer the pre-trained contextual knowledge into the monocular encoder and ensuring that the enhanced scene understanding is seamlessly integrated without compromising spatial fidelity. In general, our contributions are summarized as follows:
\begin{itemize}
    \item We design a two-stage novel weak supervision paradigm, dubbed as \textbf{CA-W3D}, which infuses context-aware semantics into monocular 3D modeling rather than relying solely on regional feature learning.

    \item We propose a \textbf{R}egion-wise \textbf{O}bject \textbf{C}ontrastive \textbf{M}atching (\textbf{ROCM}) mechanism for the initial pre-training stage, followed by a \textbf{D}ual-\textbf{to}-\textbf{O}ne \textbf{D}istillation (\textbf{D2OD}) mechanism in the subsequent pseudo-3D label training stage, enabling the monocular encoder to acquire deeper scene-level interpretative capabilities.
    
    \item Our framework demonstrates significant performance improvements on the KITTI dataset, achieving notable gains in the $AP_\text{40}$ metric and up to a 17\% improvement in $AP_\text{11}$ compared to the benchmark.
\end{itemize}

\begin{figure*}
  \centering
    \includegraphics[width=\textwidth]{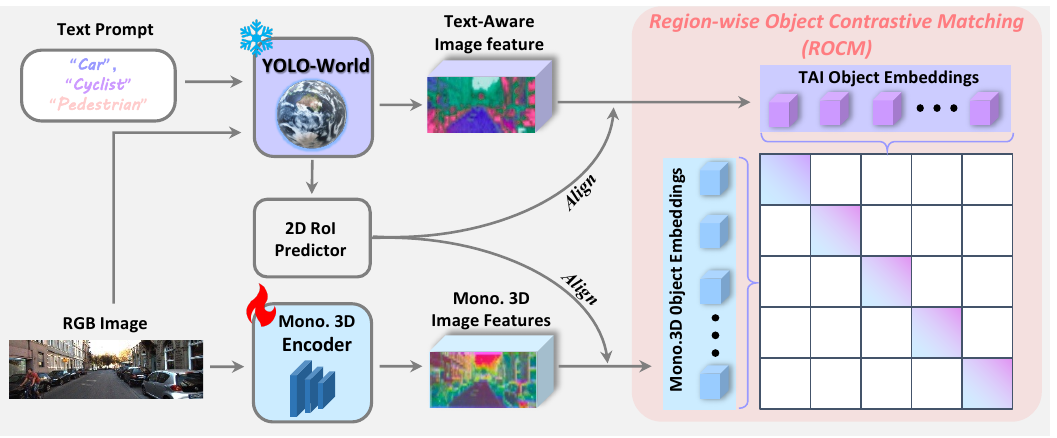} % 调整宽度
    % \includesvg[width=1\textwidth]{Figures/Framework.svg}
  \caption{\textbf{The overview of training pipeline for the pre-training stage.} Our proposed \textit{CA-W3D} framework introduces a Context-Aware Weak Supervision strategy for monocular 3D object detection. In the pre-training stage, a \textit{Region-wise Object Contrastive Matching} \textit{(ROCM)} mechanism is employed to align text-aware image features extracted from a frozen open-vocabulary detector (\textit{YOLO-World}) with monocular 3D features obtained from a trainable monocular 3D encoder. This alignment facilitates the learning of scene-specific attributes and enhances contextual reasoning.} 
  \label{fig:framework_pretrain}
  \clearpage
  \vspace{-10pt}
\end{figure*}

%% file: Related_Works.tex
\vspace{10pt}
\section{RELATED WORKS}

\subsection{\textbf{Label-Efficient Monocular 3D Detection}}
Monocular 3D object detection traditionally relies on a large volume of high-quality 3D annotations. To alleviate this dependency, researchers have explored self-supervised, weakly-supervised, which typically leverage a limited amount of labeled data or prior knowledge to infer 3D object bounding boxes from 2D image information. FGR~\cite{fgr} utilizes point cloud data and frustum-based modal features to infer 3D annotations without explicit supervision. WS3D~\cite{ws3d} employs center-based annotations and instance segmentation to refine 3D object localization. WeakM3D~\cite{peng_weakm3d_2022} leverages LiDAR point cloud priors to pre-train a monocular 3D object detector and utilizes handcrafted priors to mitigate the challenges of single-frame 3D detection. And semi-supervised methods like Casine et al. ~\cite{ zhao2024lanecmkt} proposes a knowledge distillation framework based on a Teacher-Student paradigm to enhance the learning process. While these manually designed supervision methods primarily focus on local object regions, neglecting broader semantic information. This narrow scope limits the monocular encoder’s capacity to capture contextual relationships within complex scenes, leading to a dilemma for encoder to comprehend the holistic scene context essential for robust 3D reasoning in open-world scenarios.

% SDF~\cite{sdf} integrate shape priors from DeepSDF~\cite{deepsdf} with synthetic data to reconstruct 3D structures from 2D projections.

\subsection{\textbf{Open-Vocabulary Detection}}
Open-Vocabulary Detection (OVD)~\cite{ovd} seeks to identify objects beyond a predefined set of categories. Traditional methods\cite{ovod} predominantly train from a closed set of categories and evaluating generalization to unseen classes. However, these approaches often struggle with scalability and exhibit limited adaptability when encountering entirely novel object categories. Subsequently, image-text matching techniques leverage large-scale image-text datasets to extend the detector’s vocabulary. OWL-ViTs ~\cite{OWLViTs} refines Vision Transformers (ViTs) through fine-tuning on detection and grounding datasets, enabling an efficient and streamlined OVD model. Grounding DINO~\cite{groundingDINO}  integrates grounding-based pre-training with detection transformers and cross-modal fusion techniques to enhance detection accuracy. CLIP~\cite{clip} employs contrastive learning to align textual and visual representations, facilitating robust semantic associations between textual descriptions and images. Furthermore, YOLO-World~\cite{yoloworld} incorporates region-level text-image alignment strategies to achieve better-grained recognition capabilities. These open-vocabulary detection (OVD) methods demonstrate a significant advantage in precisely aligning textual and visual representations at the regional level. This alignment facilitates a closer association between co-occurring objects within the same scene across both text and image modalities. Consequently, the monocular image encoder can achieve a higher-level understanding of overall scene composition and semantics.

%% file: Methodology_2.tex
\vspace{10pt}
\section{METHODOLOGY}
To empower the monocular encoder to recognize scene-specific object groupings while distinguishing inter-scene variations without incurring additional computational cost, we propose a pre-training schema Region-wise Object Contrastive Matching (ROCM) in Section \ref{sec:region-text object matching} and a training framework Dual-to-One Distillation (D2OD) in Section \ref{sec:double_encoder_fusion}. Besides, we introduce the loss functions employed in both the pre-training and training in Section \ref{sec:loss_function}.

\subsection{\textbf{Region-wise Object Contrastive Matching (ROCM)}}\label{sec:region-text object matching}
As shown in Fig. \ref{fig:framework_pretrain}, in this pre-training stage, our objective is to propel the monocular encoder with the capacity of differentiating scene-dependent attributes, thereby acquiring richer contextual understanding. To guide monocular image feature learning, we leverage the 2D Vision Grounding (2DVG) foundational model, YOLO-World, employing a region-text fusion strategy to generate semantic regional features as a pre-training target. These Text-Aware Object Embeddings (TAI) from YOLO-World encode higher-level contextual information through multi-modal learning, establishing a structured correspondence with monocular object embeddings, akin to the relationship between images and captions in CLIP. To capitalize on this, we introduce a novel Region-wise Object Contrastive Matching (ROCM) loss, designed to reinforce co-occurring object embeddings and enhance scene-level comprehension within the monocular encoder.

\noindent \textbf{Region and Text Information Extraction.} Given the monocular image input $I \in \mathbb{R}^{h \times w \times c}$, where $h$ and $w$ are the height and width respectively and $c$ is channel of each image. The monocular encoder is designed based on ResNet-34 to encode input into a monocular image feature $F_\text{m3d} \in \mathbb{R}^{d \times h \times w}$, where $d$ is the dimension of features. To generate high semantic object embedding as reference, we employ a frozen YOLO-World model to process image input $I$ alongside a text prompt $T \in \mathbb{R}^{n \times l}$ simultaneously, where $n$ is the number of texts and $l$ is the length of each text. This foundation model utilizes a region-language alignment module VL-PAN which contains a Text-guide CSPLayer to effectively fuse multi-modal inputs and produce a text-aware image feature $F_{tai} \in \mathbb{R}^{d \times h \times w}$. The two encoding processes are illustrated as below:
\begin{equation}\label{eq:TAI}
    \mathbf{F}_{\text{tai}} = \text{YOLO-World}(\mathbf{\textit{I}}, \mathbf{\textit{T}}), \mathbf{F}_{\text{m3d}} = \text{Mono3DEnc}(\mathbf{\textit{I}}).
\end{equation}

To generate object-wise embeddings $E_\text{m3d} \in \mathbb{R}^{n_\text{m3d\_obj} \times h \times w}$ and $E_\text{tai} \in \mathbb{R}^{n_\text{tai\_obj} \times h \times w}$ from the monocular 3D encoder and YOLO-World respectively, where  $E_\text{m3d}$ means Mono.3D Object Embeddings and $E_\text{tai}$ stands for TAI Object Embeddings, $n_\text{m3d\_obj}$ and $n_\text{tai\_obj}$ are the number of proposed Object Embeddings, we extract Regions of Interest (RoIs) from the YOLO-World detection output using a "Car" text prompt. These RoIs are then used to crop the corresponding encoded image features. To ensure consistency in activation range and channel dimensions, we employ a Multi-Layer Perceptron (MLP) for feature transformation. Notably, as TAI image features are multi-scaled, we concatenate feature maps with all resolution as contextual feature references. The process is illustrated as follows:
\begin{equation}\label{eq:roi_align}
    \begin{split}
    \mathbf{E}_{\text{m3d/tai}} = \text{RoIAlign}(\mathbf{F}_{\text{m3d/tai}}).
    \end{split}
\end{equation}

\begin{figure}[tp]
    \begin{center}
        \includegraphics[width=0.5\textwidth]{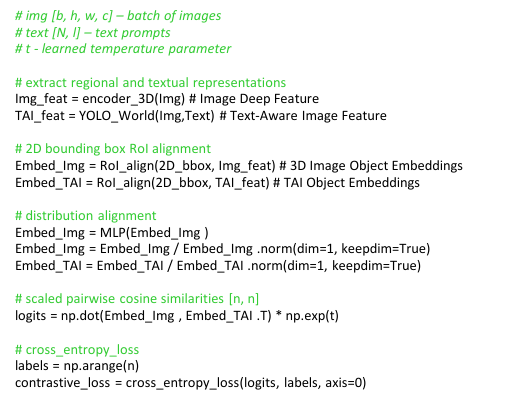}
        \caption{ 
            Numpy-like pseudocode for illustrating the Region-wise Object Conrastive Matching (ROCM) machanism.
        \label{fig:psesudo_code}}
    \end{center}
    \vspace{-15pt}
\end{figure}

\noindent \textbf{Region-wise Object Contrastive Loss.} 
Building on the aforementioned principles, we aim to establish a stronger association between co-occurring objects in the latent space while distinguishing scene-dependent attributes. Inspired by conventional contrastive loss computing for image-text matching, we extend this approach to an object-wise paradigm. This ensures that objects within a given scene share a unified global context, distinct from those in other scenes. Relying on this observation, the monocular encoder is pre-trained with deeper contextual awareness. The specific formulation of the loss function is detailed as follows:

\begin{align}\label{eq:cos_similarity}
s_{ij} &= \frac{\mathbf{E_{\text{tai}}}_i \cdot \mathbf{E_{\text{m3d}}}_j}{\|\mathbf{E_{\text{tai}}}_i\|_2 \, \|\mathbf{E_{\text{m3d}}}_j\|_2}, \quad
\mathcal{L}_i = -\log \frac{\exp(\tilde{s}_{ij})}{\sum_{k=1}^{N} \exp(\tilde{s}_{ik})}, 
\end{align}

\begin{equation}
\mathcal{L}_{\text{contrastive}} = \frac{1}{N} \sum_{i=1}^{N} \mathcal{L}_i,
\end{equation}
where $N$ is the total number of objects across all pre-defined batches and $\tilde{s}_{ij}$ is $s_{ij}$ multiplied by a temperature argument ${\tau}$. Notably, to retrain the open-vocabulary world knowledge in YOLO-World, its parameters are frozen, in some essence, the TAI embeddings provide the contextual supervision signal in this contrastive matching. The Numpy-like pseudo-code is shown in Fig. \ref{fig:psesudo_code} for clarity.

\subsection{\textbf{Dual-to-One Distillation (D2OD)}}\label{sec:double_encoder_fusion}
\begin{figure}[t]
    \begin{center}
        \includegraphics[width=0.48\textwidth]{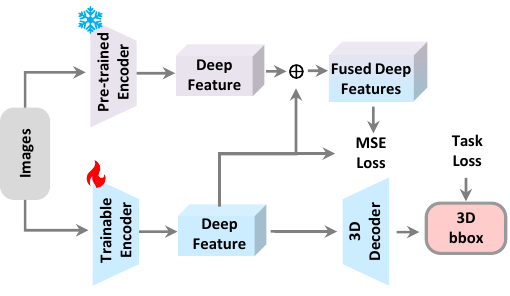}
        \caption{ 
        \textbf{Illustration of Dual-to-One Distillation (D2OD) mechanism.} The mechanism enhances monocular 3D object detection by transferring contextual knowledge from a pre-trained encoder to a trainable encoder. And a task-specific loss function is employed to optimize Mono3D-Obj detection. The proposed mechanism preserves spatial fidelity while improving contextual reasoning for monocular 3D detection. 
        \label{fig:double_encoder}}
    \end{center}
    \vspace{-20pt}
\end{figure}

We propose a Dual-to-One Distillation framework for formal training, designed to effectively transfer contextual priors into the monocular encoder while preserving spatial fidelity. As illustrated in Fig. \ref{fig:double_encoder}, the framework employs a dual-encoder architecture consisting of a context-aware pre-trained encoder ($Mono3DEnc_\text{CA}$) and a spatial trainable encoder ($Mono3DEnc_\text{SP}$), where the pre-trained encoder remains frozen to retain contextual priors. As shown in Eqs.  \ref{eq:dual_encoder}, image features $F_\text{ca} \in \mathbb{R}^{ d \times h \times w}$ and $F_\text{sp} \in \mathbb{R}^{ d \times h \times w}$ 
are extracted from pre-trained encoder and trainable encoder respectively. Then as shown in Eq. \ref{eq:concat}, these two features are concatenated and processed through a MLP to generate enhanced encoder features while maintaining the original feature dimensions: 
\begin{equation}\label{eq:dual_encoder}
\begin{aligned}
    F_{\text{\textit{ca}}} &= \operatorname{Mono3DEnc}_{\text{\textit{CA}}}(I), \\
    F_{\text{\textit{sp}}} &= \operatorname{Mono3DEnc}_{\text{\textit{SP}}}(I),
\end{aligned}
\end{equation}

\begin{equation}\label{eq:concat}
    F_{\text{\textit{fused}}} = \text{MLP}(\text{Concat}(F_{\text{\textit{ca}}}, F_{\text{\textit{sp}}})).
\end{equation}

% \begin{equation}\label{eq:concat}
% \begin{aligned}
%     F_{\text{fused}} = \operatorname{MLP}(\operatorname{concat}(F_{\text{pretrain}}, F_{\text{train}}, \text{dim}=0)).
% \end{aligned}
% \end{equation}

To ensure effective knowledge distillation, a mean squared error (MSE) loss is employed to minimize the discrepancy between the trainable encoder features and the fused features. During inference, the feature concatenation branch is no longer required. Finally, the trained encoder features are utilized for 3D bounding box regression.

\begin{table*}
    \footnotesize
    \setlength{\tabcolsep}{7pt}  % 调整列间距，默认 6pt，增大可拉宽表格
    \renewcommand{\arraystretch}{1} % 增大行高，默认 1.0
    \begin{center}
        \caption{ 
        The experimental results on the KITTI validation set for car category. 
        We can observe that our method even outperforms some fully supervised methods. 
        All the methods are evaluated with metric $AP|_{R_{11}}$, as many prior fully supervised works only provided $AP|_{R_{11}}$ results.
        }
        \label{tab:main_results}
        \begin{tabular}{l|c|cc|cc|cc}
            \toprule  
            \multirow{2}{*}{\textbf{Approaches}} & \multirow{2}{*}{\textbf{Supervision}} 
            & \multicolumn{2}{c|}{\textbf{Easy}} & \multicolumn{2}{c|}{\textbf{Moderate}} & \multicolumn{2}{c}{\textbf{Hard}} \\ 
            ~ & ~ & AP$_{BEV}$ & AP$_{3D}$ & AP$_{BEV}$ & AP$_{3D}$ & AP$_{BEV}$ & AP$_{3D}$ \\ 
            \midrule         
            Mono3D ~\cite{Mono3D} &  \multirow{7}{*}{Full}  
            & 5.22 & 2.53   &  5.19 & 2.31  &  4.13 & 2.31\\ 
            FQNet ~\cite{FQNet}  & ~ 
            & 9.50 & 5.98  & 8.02 & 5.50  & 7.71 & 4.75\\
            Deep3DBox ~\cite{Deep3DBBOX}  & ~ 
            & 9.99 & 5.85  & 7.71 & 4.10  & 5.30 & 3.84\\
            OFTNet ~\cite{OFTNet}  & ~ 
            & 11.06 & 4.07  & 8.79 & 3.27  & 8.91 & 3.29\\
            RoI-10D ~\cite{ROI10D}  & ~ 
            & 14.50 & 10.25  & 9.91 & 6.39  & 8.73 & 6.18\\
            MonoPSR ~\cite{FQNet}  & ~ 
            & 20.63 & 12.75  & 18.67 & 11.48  & 14.45 &  8.59\\ % MonoPSR
            MonoDIS \cite{AP40}   & ~  
            & 24.26 & 18.05  & 18.43 & 14.98  & 16.95 & 13.42\\ 
            M3D-RPN \cite{M3D}  & ~ 
            & \textbf{25.94} & \textbf{20.27}  & \textbf{21.18} & \textbf{17.06}  & \textbf{17.90} & \textbf{15.21}\\ 
            % Pseudo-Lidar \cite{PseudoLidar} & ~  
            % & 31.88 & 24.12  & 20.84 & 15.74  & 18.92 & 14.96\\
            % D4LCN \cite{D4LCN} & ~  
            % & 26.00 & 19.38  & 20.73 & 16.00  & 17.46 & 12.94\\
            % RTM3D \cite{RTM3D} & ~ 
            % & 25.56 & 20.77  & 22.12 & 16.86  & 20.91 & 16.63 \\
            % PatchNet \cite{PatchNet} & ~ 
            % & \bf 32.30 & 25.76  & 21.25 & 17.72  & 19.04 & 15.62\\
            % Monodle \cite{monodle} & ~ 
            % & 30.77 & 23.29  & \bf 24.53 & 20.55  & \bf 23.32 & 17.90\\
            \midrule 
            WeakM3D \cite{peng_weakm3d_2022} & \multirow{2}{*}{Weak}   
            & 24.89 & 17.06  & 16.47 & 11.63  & 14.09 & 11.17\\
            Ours & ~ 
            & \textbf{25.23} & \textbf{20.33}  & \textbf{17.89} & \textbf{13.16}  & \textbf{14.78} & \textbf{13.06}\\
            \bottomrule 
        \end{tabular}
    \end{center}
    \vspace{-5pt}
\end{table*}

% \begin{table}
% \centering
% \caption{Comparison of AP BEV/3D on KITTI validation set}
% \begin{tabular}{lccc}
% \hline
% \textbf{} & \multicolumn{3}{c}{\textbf{AP BEV / 3D}} \\
% \cline{2-4}
% \textbf{Method} & \textbf{Easy} & \textbf{Mod} & \textbf{Hard} \\
% \hline
% WeakM3D & 11.82 / 5.03 & 5.66 / 2.26 & 4.08 / 1.63 \\
% CAW3D   & 7.06 / 1.91  & 3.63 / 0.87 & 2.52 / 0.59 \\
% \hline
% \end{tabular}
% \label{tab:bev3d_results}
% \end{table}

% \begin{table}
%     \footnotesize
%     \setlength{\tabcolsep}{6pt}
%     \renewcommand{\arraystretch}{1}
%     \begin{center}
%         \caption{
%         AP$_\text{BEV}$/AP$_\text{3D}$ results on KITTI test set under the car category. 
%         }
%         \label{tab:test_result}
%         \begin{tabular}{l|cc|cc|cc}
%             \toprule
%             \multirow{2}{*}{\textbf{Method}} 
%             & \multicolumn{2}{c|}{\textbf{Easy}} 
%             & \multicolumn{2}{c|}{\textbf{Moderate}} 
%             & \multicolumn{2}{c}{\textbf{Hard}} \\
%             ~ & AP$_\text{BEV}$ & AP$_\text{3D}$ 
%             & AP$_\text{BEV}$ & AP$_\text{3D}$ 
%             & AP$_\text{BEV}$ & AP$_\text{3D}$ \\
%             \midrule
%             WeakM3D & \textbf{11.82} & \textbf{5.03} & \textbf{5.66} & \textbf{2.26} & \textbf{4.08} & \textbf{1.63} \\
%             Ours   &  7.06 & 1.91 & 3.63 & 0.87 & 2.52 & 0.59 \\
%             \bottomrule
%         \end{tabular}
%     \end{center}
%     \vspace{-20pt}
% \end{table}

% \vspace{10pt}

\subsection{\textbf{Learning Objectives}}\label{sec:loss_function}
We only use the Region-Text Object-wise Contrastive Loss for the pre-training stage. For the formal training stage, we utilize both MSE loss from Double Encoder Fusion and 3D Loss to guide model learning, where the integration of 3D pseudo label supervision during the training phase enhances detection precision, ensuring robust performance in monocular 3D object detection scenarios.

To provide 3D pseudo label for 3D loss, for each image, 3D LiDAR point clouds are projected onto its relative object 2D mask to obtain RoI points. For 3D loss computation, we directly match the RoI points with the predicted 3D bounding boxes to compute the loss. To address various challenges, we adopt key loss functions from WeakM3D~\cite{peng_weakm3d_2022} as follows.

\noindent \textbf{Geometric Alignment Loss.} As shown in Eq. \ref{eq:geometry_loss}, this loss mitigates errors in using center loss only, improving localization accuracy in 3D object detection:
\begin{equation}
\begin{aligned}
\mathcal{L}_{\text{geometry}} &= \left\| \mathbf{P} - \mathbf{P}_I \right\|_1 \\
&= \left\| \mathbf{P} - \text{Intersect} \left( \text{Ray}_{\mathbf{P}_{3d} \rightarrow \mathbf{P}},\ b_{3d} \right) \right\|_1,
\end{aligned}
\label{eq:geometry_loss}
\end{equation}
where $P$ represents RoI points, and $P_I$ represents ray (from RoI points to predicted 3D center) intersection with the edge of the 3D box prediction.

\noindent\textbf{Ray Tracing Loss.} As shown in Eq. \ref{eq:ray_tracing_loss}, this loss addresses surface uncertainty in RoI points, ensuring their precise assignment to the correct object surface:
\begin{equation}
\begin{aligned}
\mathcal{L}_{\text{ray-tracing}} = 
\begin{cases}
\left\| \mathbf{P} - \mathbf{P}_R \right\|_1, & \substack{\text{if } \text{Ray}_{\mathbf{P}_{\text{cam}} \rightarrow \mathbf{P}} \\ \text{intersects with } b_{3d}}, \\[5pt]
0, & \text{otherwise.}
\end{cases}
\end{aligned}
\label{eq:ray_tracing_loss}
\end{equation}

Suppose $P1$ and $P2$ are the intersection points of the predicted 3D bounding box's edge and the ray (from camera to the RoI point), then the $P_{R}$ is the closer point to the camera which is chosen for the loss calculation.

\noindent \textbf{Point-wise Balancing Loss.} As shown in Eq. \ref{eq:neighborhood_count}, this loss compensates for uneven points cloud distributions, preventing sparse yet critical points from being ignored, thereby enhancing overall detection robustness:
\begin{equation}
\begin{aligned}
\mathcal{E}_i = \sum_{j=1}^{M} \mathbf{1} \bigg( 
\left\| \mathbf{P}_i - \mathbf{P}_j \right\|_2 
< R \bigg),
\end{aligned}
\label{eq:neighborhood_count}
\end{equation}
where $M$ means the number of RoI points and $R$ is the threshold deciding if two points are the neighborhood.

Thus, the 3D loss can be denoted as follows: 
\begin{equation}
\mathcal{L}_{\text{3D}} = \frac{1}{M} \sum_{i=1}^{M} \left( \frac{\mathcal{L}_{\text{geometry}_i} + \mathcal{L}_{\text{ray-tracing}_i} + \lambda \mathcal{L}_{\text{center}_i}}{\mathcal{E}_i} \right),
\label{eq:balancing_loss}
\end{equation}
where $\lambda$ is a hyper parameter. 

Ultimately, we integrate the MSE loss from pre-trained contexual priors and the Mono.3D detection task loss together, where the final loss for the formal training stage is: 
\begin{equation}
\mathcal{L} = \mathcal{L}_{\text{MSE}} + \mathcal{L}_{\text{3D}}.
\label{eq:final_loss}
\end{equation}

%% file: Experiments_and_Results.tex
\section{EXPERIMENTS AND RESULTS}

We first demonstrate basic experimental setup in Section \ref{sec:experimental_setup} and implementation details in Section \ref{sec:implementation_details}, and then performance compared between our work and other previous works is shown in Section \ref{sec:main_results}. The visualization result is shown in \ref{sec:visual} We present the effectiveness of our designed Region-wise Object Contrastive Matching (ROCM) and Dual-to-One Distillation (D2OD) in Ablation Experiments in Section \ref{sec:ablation_experiments}.

\subsection{Experimental Setup}\label{sec:experimental_setup}
 We utilized the KITTI 3D Object Detection Dataset for both pre-training and training stages, which is a benchmark dataset widely used for 3D object detection and scene understanding in autonomous driving research. It consists of 7,481 labeled images for training and 7,518 unlabeled images for testing. Each labeled frame includes annotations for 3D bounding boxes of objects such as cars, pedestrians, and cyclists. We divided the dataset into a training set with 3,712 images and a validation set with 3,769 images, following a commonly used split in the community. All experiments are conducted using two NVIDIA 4090 GPUs. Consistent with the WeakM3D, we use Average Precision ($AP_\text{11}$) with three different difficulties: "\textit{Easy}", "\textit{Mod}" and "\textit{Hard}" as our evaluation metrics and both $AP_\text{11}$ and $AP_\text{40}$ for ablation study.

\subsection{Implementation Details}\label{sec:implementation_details}
\noindent \textbf{Pre-training Stage.} During the pre-training phase, the model is trained for 20 epochs with a batch size of 8 using the AdamW optimizer, set with a learning rate of $1 \times 10^{-4}$. TAI feature generation is performed using \textit{YOLO-World v1 Small} version, incorporating the CLIP model with version ViT-B-32. 

\noindent\textbf{Training Stage.} In the training phase, the model is trained for 50 epochs with a batch size of 8. The Adam optimizer is employed with a learning rate of $1 \times 10^{-4}$. The image backbone is configured using ResNet-34\cite{Resnet} to enhance feature extraction capabilities. This setup ensures a robust foundation for fine-tuning and downstream tasks.

\begin{figure*}
    \begin{center}
        \includegraphics[width=\textwidth]{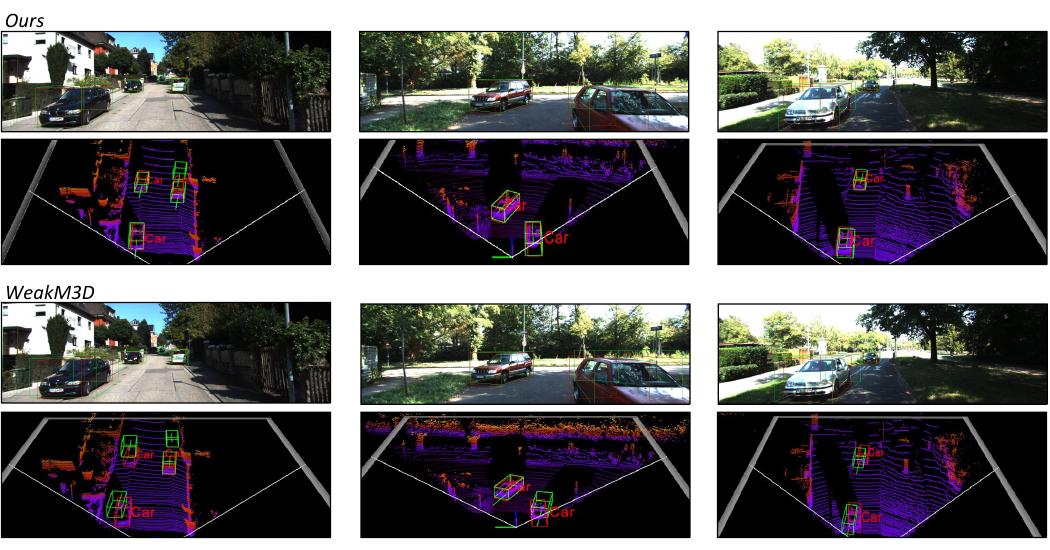}
        \caption{ 
            \textbf{Visualization comparisons between our proposed CA-W3D and WeakM3D on the KITTI validation set for the car category.} We compare the visual results with WeakM3D’s result by projected 3D bounding boxes in 2D images and Lidar point clouds. The \textcolor{green}{green} boxes are the prediction and the \textcolor{red}{red} boxes are grounding truth.
        \label{fig:Visualization}}
    \end{center}
    % \vspace{-5pt}
\end{figure*}

\subsection{Quantatative Analysis}\label{sec:main_results}
Tab. \ref{tab:main_results} compares our work with the benchmark WeakM3D and several fully-supervised methods on KITTI 3D validation set for car category. The results demonstrate that our proposed method consistently outperforms WeaM3D and even some of fully-supervised methods across almost all evaluation metrics and difficulty levels. The most significant enhancement is observed for the \textbf{ap 11 3D} metric, where the Easy category exhibits a substantial increase of 3.27. For the \textbf{ap 11 BEV} metric, the proposed approach achieves marginal improvements across all difficulty levels, particularly in the Hard category, highlighting the impact of context-aware weak supervision in monocular 3D detection. The improvements observed under the \textit{hard} setting can be attributed to the enhanced contextual knowledge model gained through our framework, which equips the model with a stronger capacity to detect co-occurring objects. Consequently, though objects are visually elusive, the model could still identify them.

% The improvements observed under the \textit{hard} setting can be attributed to that the contextual knowledge equips the model with a stronger capacity to detect co-occurring objects. Consequently, though objects are visually elusive, the model could still identify them. While, Table~\ref{tab:test_result} highlights the limited generalization of our method on the KITTI test set.

\subsection{Qualitative Analysis}\label{sec:visual}
Fig. \ref{fig:Visualization} illustrates the 3D object detection results of our method and our benchmark WeakM3D on the KITTI dataset. The upper two rows present our 3D detection results in RGB images and corresponding LiDAR point. The lower two rows show the result of WeakM3D in these two modalities. Each column corresponds to a distinct scene, where green annotations denote the predicted 3D bounding boxes, and red annotations represent the ground truth. As observed, our method demonstrates superior detection accuracy compared to the benchmark approach. Particularly for closer objects with minimal occlusion, our method excels by leveraging larger, more complete objects as high-quality learning samples for ROCM. This enhances target feature alignment and contextual prior knowledge acquisition, by which the model gains stronger supervision signals to improve scene-specific learning.

\subsection{Ablation Experiments}\label{sec:ablation_experiments}

% 实验目的
% 实验内容/结果
% 结果分析/原因
% 实验结论
\noindent \textbf{Effects of Alignment Strategy.} We investigated the effectiveness of the proposed Region-wise Object Contrastive Matching (ROCM) in aligning TAI Object Embeddings with Mono.3D Image Embeddings. Tab. \ref{tab:alignment_strategy} demonstrates that the contrastive loss offers significant advantages over conventional alignment strategies, such as Mean Squared Error (MSE) Loss and Kullback-Leibler (KL) Divergence Loss~\cite{kl_loss}, where contrastive loss is about 3 AP higher than KL Loss and 2 AP higher than MSE Loss. This superior performance can be attributed to the contrastive loss’s ability to minimize discrepancies between similar co-occurrence object pairs while maximizing the distinction between dissimilar pairs, thereby facilitating more accurate alignment. In contrast, MSE and KL losses treat all objects holistically, focusing on global differences rather than fine-grained pairwise relationships, which limits their effectiveness in object-level alignment tasks.
\begin{table}[t]
\caption{\textbf{Ablations on Alignment Strategies.} We compare the effects of different matching loss for object embedding in the first pre-training stage.}
\label{tab:alignment_strategy}
\centering

\begin{subtable}[t]{\columnwidth}
\centering
\caption{Tested on $R$=40 and $IoU$=0.5}
\resizebox{\columnwidth}{!}{ % 让表格适应单栏宽度
\begin{tabular}{l|ccc|ccc}
\toprule
\multirow{2}{*}{Loss Function} & \multicolumn{3}{c|}{AP$_{BEV}$} & \multicolumn{3}{c}{AP$_{3D}$} \\
% \cline{2-7}
 & \textit{Easy} & \textit{Mod} & \textit{Hard} & \textit{Easy} & \textit{Mod} & \textit{Hard} \\
\midrule
\makecell{Kullback-Leibler Divergence (KL)} & 42.57 & 28.00 & 20.98 & 35.89 & 22.01 & 16.78 \\
Mean Squared Error (MSE) & 43.49 & 28.74 & 22.70 & 37.21 & 22.82 & 18.01 \\
\textbf{ROCM (Ours)} & \textbf{45.63} & \textbf{29.36} & \textbf{23.88} & \textbf{39.11} & \textbf{23.54} & \textbf{18.96} \\
\bottomrule
\end{tabular}
}
\label{tab:ap40_alignment}
\end{subtable}

\vspace{0.3cm}

\begin{subtable}[t]{\columnwidth} % 适应单栏
\centering
\caption{Tested on $R$=11 and $IoU$=0.7}
\label{tab:ap11_alignment} % 确保 label 放在 caption 之后
\resizebox{\columnwidth}{!}{ % 适配单栏
\begin{tabular}{l|ccc|ccc}
\toprule
\multirow{2}{*}{Loss Function} & \multicolumn{3}{c|}{AP$_{BEV}$} & \multicolumn{3}{c}{AP$_{3D}$} \\
 & \textit{Easy} & \textit{Mod} & \textit{Hard} & \textit{Easy} & \textit{Mod} & \textit{Hard} \\
\midrule
\makecell{Kullback-Leibler Divergence (KL)} & 16.50 & 12.59 & 8.01 & 10.87 & 9.09 & 9.09 \\
Mean Squared Error (MSE) & 16.35 & 12.98 & 10.98 & 11.56 & 10.22 & 9.09 \\
\textbf{ROCM (Ours)} & \textbf{19.88} & \textbf{14.95} & \textbf{12.98} & \textbf{14.13} & \textbf{11.23} & \textbf{10.57} \\
\bottomrule
\end{tabular}
}
\end{subtable}

% \begin{subtable}[t]{\linewidth}
% \centering
% \caption{Tested on $R$=11 and $IoU$=0.7}
% \begin{tabular}{l|ccc|ccc}
% \toprule
% \multirow{2}{*}{Loss Function} & \multicolumn{3}{c|}{AP$_{BEV}$} & \multicolumn{3}{c}{AP$_{3D}$} \\
% % \cline{2-7}
%  & \textit{Easy} & \textit{Mod} & \textit{Hard} & \textit{Easy} & \textit{Mod} & \textit{Hard} \\
% \midrule
% Kullback-Leibler Divergence (KL) & 16.50 & 12.59 & 8.01 & 10.87 & 9.09 & 9.09 \\
% Mean Squared Error (MSE) & 16.35 & 12.98 & 10.98 & 11.56 & 10.22 & 9.09 \\
% \textbf{Ours} & \textbf{19.88} & \textbf{14.95} & \textbf{12.98} & \textbf{14.13} & \textbf{11.23} & \textbf{10.57} \\
% \bottomrule
% \end{tabular}
% \label{tab:ap11_alignment}
% \end{subtable}
\vspace{-5pt}
\end{table}

\noindent\textbf{Effects of Pre-trained Encoder Settings.} This experiment explores the effectiveness of integrating knowledge from a pre-trained encoder with an initial Mono.3D encoder to guide formal training. We designed four kinds of utilities as follows: 
\begin{itemize}
\item \textit{Single-Stage Training.} Training is conducted without any pre-training phase. The contrastive loss is directly combined with the task loss during the optimization process, allowing both losses to guide the training simultaneously.

\item \textit{Two-Stage Training.} Two distinct phases are involved. In the first stage, the encoder undergoes pre-training using contrastive loss. In the second stage, the pre-trained encoder is loaded, and formal training is carried out using only the task loss.

\item \textit{Double Encoder Distillation.} The model is first pre-trained with contrastive loss. During formal training, both the training and pre-trained encoders are loaded, and MSE loss is computed between their features. This MSE loss is combined with the task loss to improve performance.

\item \textit{Double Encoder Fusion.} Similar to Double Encoder Distillation, but with an added fusion step. Pre-trained and trainable encoder features are fused before computing MSE loss, ensuring comparison with the enhanced representation.
\end{itemize}

As shown in Tab. \ref{tab:pretrained_encoder_usage}, the Double Encoder Fusion approach consistently outperforms the other three methods. Compared to the 1-stage method, since contrastive loss and task loss optimize different objectives, combining them directly in early training stages can lead to conflicts, potentially disrupting convergence and harming overall performance. As to the 2-stage method, the spatial features learned using contrastive loss during the pre-training stage may not fully align with the requirements of the downstream task detection task, leading to suboptimal transfer of learned knowledge during formal training. Feature fusion in the Double Encoder Fusion method creates richer and more comprehensive feature representations, enabling the model to capture finer-grained details. This enhanced representation ultimately boosts the model’s ability to extract meaningful patterns, improving overall performance.
\begin{table}[htbp]
\caption{\textbf{Ablations on Pre-trained Encoder Settings}. 1-stage stands for directly combining the contrastive loss with the task loss and there is only 1 training stage. 2-stage means contrastive loss for pre-training and task loss for formal training. Double encoder refers to loading both pre-trained and initialized encoders during training, where distillation calculates the loss between these encoders directly, and fusion requires feature fusion before loss calculation.}
\label{tab:pretrained_encoder_usage}
\centering

\begin{subtable}[t]{\linewidth}
\centering
\caption{Tested on $R$=40 and $IoU$=0.5}
\resizebox{\linewidth}{!}{
\begin{tabular}{l|ccc|ccc}
\toprule
\multirow{2}{*}{Pre-trained Encoder Settings} & \multicolumn{3}{c|}{AP$_{BEV}$} & \multicolumn{3}{c}{AP$_{3D}$} \\
 & \textit{Easy} & \textit{Mod} & \textit{Hard} & \textit{Easy} & \textit{Mod} & \textit{Hard} \\
\midrule
1-stage & 34.86 & 18.50 & 14.03 & 29.03 & 14.48 & 11.35 \\
2-stage & 38.97 & 24.66 & 19.79 & 32.80 & 19.74 & 15.44 \\
Double Encoder Distillation & 44.41 & 27.72 & 22.22 & 37.70 & 22.34 & 17.61 \\
\textbf{D2OD (Ours)} & \textbf{45.63} & \textbf{29.36} & \textbf{23.88} & \textbf{39.11} & \textbf{23.54} & \textbf{18.96} \\
\bottomrule
\end{tabular}
}
\label{tab:ap40_encoder}
\end{subtable}

\vspace{0.3cm}

\begin{subtable}[t]{\linewidth}
\centering
\caption{Tested on $R$=11 and $IoU$=0.7}
\resizebox{\linewidth}{!}{
\begin{tabular}{l|ccc|ccc}
\toprule
\multirow{2}{*}{Pre-trained Encoder Settings} & \multicolumn{3}{c|}{AP$_{BEV}$} & \multicolumn{3}{c}{AP$_{3D}$} \\
 & \textit{Easy} & \textit{Mod} & \textit{Hard} & \textit{Easy} & \textit{Mod} & \textit{Hard} \\
\midrule
1-stage & 15.18 & 11.27 & 10.97 & 11.16 & 9.61 & 9.09 \\
2-stage & 15.20 & 12.16 & 11.22 & 11.59 & 10.28 & 9.61 \\
Double Encoder Distillation & 15.15 & 12.16 & 11.19 & 10.64 & 9.62 & 9.47 \\
\textbf{D2OD (Ours)} & \textbf{19.88} & \textbf{14.95} & \textbf{12.98} & \textbf{14.13} & \textbf{11.23} & \textbf{10.57} \\
\bottomrule
\end{tabular}
}
\label{tab:ap11_encoder}
\end{subtable}

\end{table}

\begin{table}[htbp]
\caption{\textbf{Ablations on Fusion Strategies.} Performance of three fusion strategies to integrate features from the pre-trained encoder and trainable encoder.}
\label{tab:fusion_strategy}
\centering

\begin{subtable}[t]{\linewidth}
\centering
\caption{Tested on $R$=40 and $IoU$=0.5}
\resizebox{\linewidth}{!}{
\begin{tabular}{l|ccc|ccc}
\toprule
\multirow{2}{*}{Fusion Strategy} & \multicolumn{3}{c|}{AP$_{BEV}$} & \multicolumn{3}{c}{AP $_{3D}$} \\
 &\textit{Easy} & \textit{Mod} & \textit{Hard} & \textit{Easy} & \textit{Mod} & \textit{Hard} \\
\midrule
Cross Attention \textbf{w/} FNN & 43.59 & 28.67 & 22.60 & 36.34 & 22.01 & 17.30 \\
Cross Attention \textbf{w/o} FNN & 44.59 & \textbf{29.79} & \textbf{24.35} & 38.18 & 23.01 & 18.43 \\
\textbf{Concatenation} & \textbf{45.63} & 29.36 & 23.88 & \textbf{39.11} & \textbf{23.54} & \textbf{18.96} \\
\bottomrule
\end{tabular}
}
\label{tab:ap40_fusion}
\end{subtable}

\vspace{0.3cm} % 子表格间距

\begin{subtable}[t]{\linewidth}
\centering
\caption{Tested on $R$=11 and $IoU$=0.7}
\resizebox{\linewidth}{!}{
\begin{tabular}{l|ccc|ccc}
\toprule
\multirow{2}{*}{Fusion Strategy} & \multicolumn{3}{c|}{AP$_{BEV}$} & \multicolumn{3}{c}{AP $_{3D}$} \\
 & \textit{Easy} & \textit{Mod} & \textit{Hard} & \textit{Easy} & \textit{Mod} & \textit{Hard} \\
\midrule
Cross Attention \textbf{w/} FNN & 18.57 & 14.23 & 11.90 & 13.48 & 10.88 & 10.10 \\
Cross Attention \textbf{w/o} FNN & 18.59 & 13.85 & 11.89 & 13.35 & 10.74 & 9.72 \\
\textbf{Concatenation} & \textbf{19.88} & \textbf{14.95} & \textbf{12.98} & \textbf{14.13} & \textbf{11.12} & \textbf{10.57} \\
\bottomrule
\end{tabular}
}
\label{tab:ap11_fusion}
\end{subtable}
\vspace{-15pt}
\end{table}

\noindent \textbf{Effects of Fusion Strategies.} This experiment validates the effectiveness of various fusion strategies for integrating pre-trained encoder features with trainable encoder features of Double Encoder Fusion strategy. Tab. \ref{tab:fusion_strategy} presents that the Concatenation method offers superior performance compared to the other two approaches: Cross Attention without Feed-Forward Network (FNN) and Cross Attention with FNN. The observed performance differences can be attributed to the characteristics of the KITTI 3D Object Detection dataset, which is relatively small and may not provide sufficient trainable samples or time for the model to converge fully. Consequently, the Cross Attention mechanisms may fail to adequately distribute attention weights, leading to suboptimal feature integration. In contrast, the Concatenation strategy offers a more straightforward and stable fusion method, effectively preserving and combining information from both pre-trained and trainable encoder features. This results in improved feature representation and overall model performance.

%% file: Conclusion_and_Discussion.tex
\vspace{10pt}
\section{CONCLUSION}

In this paper, we introduced a novel Context-Aware Weak Supervision framework for monocular 3D object detection to address the limitations of conventional weakly supervised methods in capturing global scene context. We design a two-stage training paradigm: Region-wise Object Contrastive Matching (ROCM), which aligns regional visual and text knowledge to enhance contextual understanding, and Dual-to-One Distillation (D2OD) to effectively transfer this knowledge to the monocular encoder while preserving spatial fidelity. Evaluations on the KITTI benchmark demonstrate a significantly improves detection performance. Offering a more robust and generalizable solution for real-world autonomous perception. However, the current text prompts in Stage 1 (ROCM) are limited to single-word descriptions. For the future work, we will explore more fine-grained and diverse textual expressions to improve regional visual-text alignment.

\vspace{30pt}

%% file: paper.bbl
\begin{thebibliography}{10}
\providecommand{\url}[1]{#1}
\csname url@rmstyle\endcsname
\providecommand{\newblock}{\relax}
\providecommand{\bibinfo}[2]{#2}
\providecommand\BIBentrySTDinterwordspacing{\spaceskip=0pt\relax}
\providecommand\BIBentryALTinterwordstretchfactor{4}
\providecommand\BIBentryALTinterwordspacing{\spaceskip=\fontdimen2\font plus
\BIBentryALTinterwordstretchfactor\fontdimen3\font minus \fontdimen4\font\relax}
\providecommand\BIBforeignlanguage[2]{{%
\expandafter\ifx\csname l@#1\endcsname\relax
\typeout{** WARNING: IEEEtran.bst: No hyphenation pattern has been}%
\typeout{** loaded for the language `#1'. Using the pattern for}%
\typeout{** the default language instead.}%
\else
\language=\csname l@#1\endcsname
\fi
#2}}

\bibitem{xia2024rgbd}
H.~Xia, Y.~Fu, S.~Liu, and X.~Wang, ``Rgbd objects in the wild: scaling real-world 3d object learning from rgb-d videos,'' in \emph{CVPR}, 2024, pp. 22\,378--22\,389.

\bibitem{pesavento2024anim}
M.~Pesavento, Y.~Xu, N.~Sarafianos, R.~Maier, Z.~Wang, C.-H. Yao, M.~Volino, E.~Boyer, A.~Hilton, and T.~Tung, ``Anim: Accurate neural implicit model for human reconstruction from a single rgb-d image,'' in \emph{CVPR}, 2024, pp. 5448--5458.

\bibitem{huang2024matchu}
J.~Huang, H.~Yu, K.-T. Yu, N.~Navab, S.~Ilic, and B.~Busam, ``Matchu: Matching unseen objects for 6d pose estimation from rgb-d images,'' in \emph{CVPR}, 2024, pp. 10\,095--10\,105.

\bibitem{lang2019pointpillars}
A.~H. Lang, S.~Vora, H.~Caesar, L.~Zhou, J.~Yang, and O.~Beijbom, ``Pointpillars: Fast encoders for object detection from point clouds,'' in \emph{CVPR}, 2019, pp. 12\,697--12\,705.

\bibitem{zhou2017voxelnetendtoendlearningpoint}
Y.~Zhou and O.~Tuzel, ``Voxelnet: End-to-end learning for point cloud based 3d object detection,'' \emph{arXiv preprint arXiv:1711.06396}, 2017.

\bibitem{zhao2024advancements}
R.~Zhao, Y.~Heng, H.~Wang, Y.~Gao, S.~Liu, C.~Yao, J.~Chen, and W.~Cai, ``Advancements in 3d lane detection using lidar point clouds: From data collection to model development,'' in \emph{ICRA}.\hskip 1em plus 0.5em minus 0.4em\relax IEEE, 2024, pp. 5382--5388.

\bibitem{yu2024unleashing}
R.~Yu, R.~Zhao, J.~Li, Q.~Zhao, S.~Zhu, H.~Yan, and M.~Wang, ``Unleashing the potential of mamba: Boosting a lidar 3d sparse detector by using cross-model knowledge distillation,'' \emph{arXiv preprint arXiv:2409.11018}, 2024.

\bibitem{yu2024future}
R.~Yu, R.~Zhao, C.~Nie, H.~Wang, H.~Yan, and M.~Wang, ``Future does matter: Boosting 3d object detection with temporal motion estimation in point cloud sequences,'' \emph{arXiv preprint arXiv:2409.04390}, 2024.

\bibitem{zhang2023odm3d}
W.~Zhang, D.~Liu, C.~Ma, and W.~Cai, ``{ODM3D: Alleviating Foreground Sparsity for Semi-Supervised Monocular 3D Object Detection},'' in \emph{WACV}, 2024, pp. 7542--7552.

\bibitem{weng2019monocular}
X.~Weng and K.~Kitani, ``Monocular 3d object detection with pseudo-lidar point cloud,'' in \emph{CVPRW}, 2019, pp. 0--0.

\bibitem{wang_plumenet_2021}
Y.~Wang, B.~Yang, R.~Hu, M.~Liang, and R.~Urtasun, ``{PLUMENet}: {Efficient} {3D} {Object} {Detection} from {Stereo} {Images},'' in \emph{IROS}, Sept. 2021, pp. 3383--3390, iSSN: 2153-0866.

\bibitem{you_pseudo-lidar_2020}
Y.~You, Y.~Wang, W.-L. Chao, D.~Garg, G.~Pleiss, B.~Hariharan, M.~Campbell, and K.~Q. Weinberger, ``Pseudo-lidar++: Accurate depth for 3d object detection in autonomous driving,'' \emph{arXiv preprint arXiv:1906.06310}, 2020.

\bibitem{wang_pseudo-lidar_2019}
Y.~Wang, W.-L. Chao, D.~Garg, B.~Hariharan, M.~Campbell, and K.~Q. Weinberger, ``\BIBforeignlanguage{en}{Pseudo-{LiDAR} {From} {Visual} {Depth} {Estimation}: {Bridging} the {Gap} in {3D} {Object} {Detection} for {Autonomous} {Driving}},'' in \emph{\BIBforeignlanguage{en}{CVPR}}.\hskip 1em plus 0.5em minus 0.4em\relax Long Beach, CA, USA: IEEE, June 2019, pp. 8437--8445.

\bibitem{qian_end--end_2020}
R.~Qian, D.~Garg, Y.~Wang, Y.~You, S.~Belongie, B.~Hariharan, M.~Campbell, K.~Q. Weinberger, and W.-L. Chao, ``End-to-end pseudo-lidar for image-based 3d object detection,'' in \emph{CVPR}, 2020, pp. 5881--5890.

\bibitem{leonardis_weakly_2025}
K.-C. Huang, Y.-H. Tsai, and M.-H. Yang, ``Weakly supervised 3d object detection via multi-level visual guidance,'' in \emph{ECCV}, 2024, pp. 175--191.

\bibitem{jiang_weakly_2024}
X.~Jiang, S.~Jin, L.~Lu, X.~Zhang, and S.~Lu, ``Weakly supervised monocular 3d detection with a single-view image,'' in \emph{CVPR}, 2024, pp. 10\,508--10\,518.

\bibitem{zhang_decoupled_nodate}
J.~Zhang, J.~Li, X.~Lin, W.~Zhang, X.~Tan, J.~Han, E.~Ding, J.~Wang, and G.~Li, ``Decoupled pseudo-labeling for semi-supervised monocular 3d object detection,'' in \emph{CVPR}, 2024, pp. 16\,923--16\,932.

\bibitem{peng_weakm3d_2022}
L.~Peng, S.~Yan, B.~Wu, Z.~Yang, X.~He, and D.~Cai, ``Weakm3d: Towards weakly supervised monocular 3d object detection,'' in \emph{ICLR}, 2022.

\bibitem{oquab2024dinov2learningrobustvisual}
M.~Oquab, T.~Darcet, and M.~et. al., ``{DINOv2}: Learning robust visual features without supervision,'' \emph{arXiv preprint arXiv:2304.07193}, 2023.

\bibitem{singh2024smartmask}
J.~Singh, J.~Zhang, Q.~Liu, C.~Smith, Z.~Lin, and L.~Zheng, ``Smartmask: context aware high-fidelity mask generation for fine-grained object insertion and layout control,'' in \emph{CVPR}, 2024, pp. 6497--6506.

\bibitem{radford2021learningtransferablevisualmodels}
A.~Radford, J.~W. Kim, C.~Hallacy, A.~Ramesh, G.~Goh, S.~Agarwal, G.~Sastry, A.~Askell, P.~Mishkin, J.~Clark, G.~Krueger, and I.~Sutskever, ``Learning transferable visual models from natural language supervision,'' \emph{arXiv preprint arXiv:2103.00020}, 2021.

\bibitem{fgr}
Y.~Wei, S.~Su, J.~Lu, and J.~Zhou, ``Fgr: Frustum-aware geometric reasoning for weakly supervised 3d vehicle detection,'' in \emph{ICRA}, 2021, pp. 4348--4354.

\bibitem{ws3d}
Q.~Meng, W.~Wang, T.~Zhou, J.~Shen, L.~Van~Gool, and D.~Dai, ``Weakly supervised 3d object detection from lidar point cloud,'' in \emph{ECCV}, 2020, pp. 515--531.

\bibitem{zhao2024lanecmkt}
R.~Zhao, H.~Wang, and W.~Cai, ``Lanecmkt: Boosting monocular 3d lane detection with cross-modal knowledge transfer,'' in \emph{ACM MM}, 2024, pp. 4283--4291.

\bibitem{ovd}
A.~Zareian, K.~Dela~Rosa, D.~H. Hu, and S.-F. Chang, ``Open-vocabulary object detection using captions,'' in \emph{CVPR}, 2021, pp. 14\,393--14\,402.

\bibitem{ovod}
X.~Gu, T.-Y. Lin, W.~Kuo, and Y.~Cui, ``Open-vocabulary object detection via vision and language knowledge distillation,'' in \emph{ICLR}, 2022.

\bibitem{OWLViTs}
M.~Minderer, A.~A. Gritsenko, A.~Stone, M.~Neumann, D.~Weissenborn, A.~Dosovitskiy, A.~Mahendran, A.~Arnab, M.~Dehghani, Z.~Shen, X.~Wang, X.~Zhai, T.~Kipf, and N.~Houlsby, ``Simple open-vocabulary object detection with vision transformers,'' in \emph{ECCV}, 2022.

\bibitem{groundingDINO}
S.~Liu, Z.~Zeng, T.~Ren, F.~Li, H.~Zhang, J.~Yang, Q.~Jiang, C.~Li, J.~Yang, H.~Su, \emph{et~al.}, ``Grounding dino: Marrying dino with grounded pre-training for open-set object detection,'' in \emph{ECCV}, 2024, pp. 38--55.

\bibitem{clip}
K.~Zhou and P.~Krähenbühl, ``Probabilistic 3d surface reconstruction from noisy multi-view depth images,'' \emph{arXiv preprint arXiv:2103.00020}, 2021.

\bibitem{yoloworld}
T.~Cheng, L.~Song, Y.~Ge, W.~Liu, X.~Wang, and Y.~Shan, ``Yolo-world: Real-time open-vocabulary object detection,'' in \emph{CVPR}, 2024, pp. 16\,901--16\,911.

\bibitem{Mono3D}
X.~Chen, K.~Kundu, Z.~Zhang, H.~Ma, S.~Fidler, and R.~Urtasun, ``Monocular 3d object detection for autonomous driving,'' in \emph{CVPR}, 2016, pp. 2147--2156.

\bibitem{FQNet}
L.~Liu, J.~Lu, C.~Xu, Q.~Tian, and J.~Zhou, ``Deep fitting degree scoring network for monocular 3d object detection,'' in \emph{CVPR}, 2019, pp. 1057--1066.

\bibitem{Deep3DBBOX}
A.~Mousavian, D.~Anguelov, J.~Flynn, and J.~Kosecka, ``3d bounding box estimation using deep learning and geometry,'' in \emph{CVPR}, 2017, pp. 7074--7082.

\bibitem{OFTNet}
T.~Roddick, A.~Kendall, and R.~Cipolla, ``Orthographic feature transform for monocular 3d object detection,'' \emph{arXiv preprint arXiv:1811.08188}, 2018.

\bibitem{ROI10D}
F.~Manhardt, W.~Kehl, and A.~Gaidon, ``Roi-10d: Monocular lifting of 2d detection to 6d pose and metric shape,'' in \emph{CVPR}, 2019, pp. 2069--2078.

\bibitem{AP40}
A.~Simonelli, S.~R. Bulo, L.~Porzi, M.~L{\'o}pez-Antequera, and P.~Kontschieder, ``Disentangling monocular 3d object detection,'' in \emph{CVPR}, 2019, pp. 1991--1999.

\bibitem{M3D}
G.~Brazil and X.~Liu, ``M3d-rpn: Monocular 3d region proposal network for object detection,'' in \emph{CVPR}, 2019, pp. 9287--9296.

\bibitem{Resnet}
K.~He, X.~Zhang, S.~Ren, and J.~Sun, ``Deep residual learning for image recognition,'' in \emph{CVPR}, 2016, pp. 770--778.

\bibitem{kl_loss}
S.~Kullback and R.~A. Leibler, ``On information and sufficiency,'' \emph{The Annals of Mathematical Statistics}, vol.~22, no.~1, pp. 79--86, 1951.

\end{thebibliography}
